\documentclass{llncs}
\usepackage{makeidx}  % allows for index generation
\usepackage{amsmath}
\usepackage{amsfonts}
\usepackage{amscd}
\usepackage[tableposition=top]{caption}
\usepackage{ifthen}
\usepackage[utf8]{inputenc}
\usepackage{graphicx}
\usepackage{graphicx}
\usepackage{caption}
\usepackage{subcaption}
\usepackage{color}

\usepackage{makeidx}
\usepackage{verbatim}
\usepackage{placeins}

\usepackage{amssymb}
\usepackage{bbm}

\usepackage{multirow} 
\usepackage{makecell}
\usepackage{hhline} 
\usepackage{array} 

\usepackage{siunitx}

\usepackage{dcolumn}
\newcolumntype{.}{D{.}{.}{-1}}

\begin{document}
\frontmatter          % for the preliminaries
\pagestyle{headings}  % switches on printing of running heads
\addtocmark{Active Learning} % additional mark in the TOC
\mainmatter              % start of the contributions
\title{When does Active Learning Work?}
\titlerunning{When does Active Learning Work?}  % abbreviated title (for running head)
\author{
Lewis P. G. Evans\inst{1} \and Niall M. Adams \inst{1} \inst{2} \and Christoforos Anagnostopoulos \inst{1}
}
\authorrunning{Lewis P. G. Evans et al.} % abbreviated author list (for running head)
\tocauthor{Lewis P. G. Evans, Niall M. Adams and Christoforos Anagnostopoulos}
\institute{Department of Mathematics, Imperial College London\\
%\email{Lewis.Evans10@imperial.ac.uk}\\ 
%WWW home page: \texttt{http://www2.imperial.ac.uk/\textasciitilde LPE10/}
\and
Heilbronn Institute for Mathematical Research, University of Bristol
}

\maketitle              % typeset the title of the contribution

\begin{abstract}
%Classification is an important area of statistical inference and machine learning. 
%Within classification, Active Learning (AL) methods seek to improve classifier performance when labels are expensive or scarce.
Active Learning (AL) methods seek to improve classifier performance when labels are expensive or scarce.
We consider two central questions: Where does AL work? How much does it help?
%To address these questions, we present a comprehensive experimental simulation study of Active Learning. 
To address these questions, a comprehensive experimental simulation study of Active Learning is presented. 
%We present a comprehensive experimental survey of Active Learning. 
%This survey addresses two questions: Where does AL work? How much does it help?
%%This survey addresses the questions of does AL work, and where AL is effective. 
We consider a variety of tasks, classifiers and other AL factors, to present a broad exploration of AL performance in various settings.
A precise way to quantify performance is needed in order to know when AL works.
%We need a precise way to quantify performance, in order to know when AL works.
%In order to know when AL works, we need a precise way to quantify performance.
Thus we also present a detailed methodology for tackling the complexities of assessing AL performance in the context of this experimental study.

\end{abstract}
\section{Introduction}
%

%%%\textcolor{red}{red_text_1} % no joy
%{\color{red} some red text}
%{\color{blue} some blue text}

Active Learning (AL) is an important sub-field of classification, where a learning system can intelligently select unlabelled examples for labelling, to improve classifier performance.
The need for AL is often motivated by practical concerns: labelled data is often scarce or expensive compared to unlabelled data \cite{Settles2009}.
%Active Learning also promises to shed light on classification itself through the questions of how and why AL works.

We consider two central questions: Where does AL work? How much does it help?
These questions are as yet unresolved, and answers would enable researchers to tackle the subsequent questions of how and why AL works.
%Those two questions: why matter; how answer, why that answer, results. THIS IS C3.
% ie: make explicit the links; NA, "this also links throughout the doc"

Several studies have shown that it is surprisingly difficult for AL to outperform the simple benchmark of random selection (\cite{Cawley2011,Provost2010}).
%Several studies have shown that for AL to outperform random selection is difficult (\cite{Cawley2011,Provost2010}).
Further, both AL methods and random selection often show high variability which makes comparisons difficult.
There are many studies showing positive results, for example \cite{Settles2009,Guyon2011}.
Notably there are several studies showing negative results, for example \cite{Baldridge2004,Provost2010}.
%However, these studies describe results in small localised regions of AL factor space (for example, a single classifier and task, and a %fixed set of AL factors).
While valuable, such studies do not permit any overview of where and how much AL works. %, as a function of the AL factors.
%%While valuable, such studies do not permit any overview of the AL experimental result surface.
Moreover, this contradiction suggests there are still things to understand, which is the objective of this paper.

We take the view that a broader study should try to understand which factors might be expected to affect AL performance. 
Such factors include the classification task and the classifier; see Section 2.3.
%Some factors that might be expected to affect AL performance include the classification task and the classifier.
%, see Section \ref{Active_Learning_Factors_Marker}.
We present a comprehensive simulation study of AL, where many AL factors are systematically varied and subsequently subjected to statistical analysis.
%We present a comprehensive experimental survey of Active Learning, where many AL factors are systematically varied.
%Those factors include task and classifier, see Section \ref{Active_Learning_Factors_Marker}.

%Those factors include task, classifier, task difficulty, task feature vector discretisation, and quantity of initially labelled data, see Section %\ref{Active_Learning_Factors_Marker}.

%This survey addresses two questions: Where does AL work? How much does it help?
%Those two questions: why, how answer, why that answer, results.
%%This survey addresses two questions: Does AL work? Where is it effective?
%%These two questions are essential for the later questions of how and why AL works.

%We use this survey to address two central questions: Where does AL work? How much does it help?
%These questions matter because they are as yet unresolved, and also because the answers would enable researchers to tackle the later %questions of how and why AL works.
%Those two questions: why matter; how answer, why that answer, results. THIS IS C3.
% ie: make explicit the links; NA, "this also links throughout the doc"

%There have been a few experimental surveys of AL, e.g. \cite{Guyon2011}.
%However there are several important methodological subtleties with the evaluation of AL performance.
%There are several important methodological subtleties with the evaluation of AL performance.
Careful reasoning about the design of AL experiments raises a number of important methodological issues with the evaluation of AL performance.
This paper contributes an assessment metholodology in the context of simulation studies to address those issues.
%in Section \ref{Methodology_Marker}.
%This paper contributes an assessment metholodology to fully address those complexities, in order to give an accurate answer to the %questions of where and how much AL works; see Section \ref{Methodology_Marker}.

For practical applications of AL, there is usually no holdout test dataset with which to assess performance.
That creates major unresolved difficulties, for example the inability to assess AL method performance, as discussed in \cite{Provost2010}.
Hence this study focusses on simulated data, so that AL performance can be assessed.
%For this we focus on simulated data, so that AL performance can be assessed.
%There are major unresolved difficulties in the practical applications of AL, many related to the absence of a holdout test dataset with which %to assess performance, see \cite{Provost2010}.
%Those difficulties are not addressed here; for this survey we use simulated data.

The structure of this paper is as follows: we present background on classification and AL in Section \ref{Background_Marker}, then describe the experimental method and assessment methodology in Sections \ref{Experimental_Method_Marker} and \ref{Methodology_Marker}.
Finally we present results in Section \ref{Results_Marker} and conclude in Section \ref{Conclusion_Marker}.

% note on syntax for S_i, delta-S_i, A_i,
%$ ({S_i}) $
%$ (S_i) $
%$ \boldsymbol{S_i} $

\section{Background}
\label{Background_Marker}

This section presents the more detailed background on classification and AL.
%This section presents the two essential background topics of classification and Active Learning in more detail.
%(Note, two subsections removed for IDA page limit, will restore them for final submission).

%
\subsection{Classification}
%

%The task of classification is to correctly predict the true classes of examples using their properties.
%%To simplify heavily, classification is regression with a discrete categorical response.
%In a sense, classification is regression with a discrete categorical response.
%The goal of classification is to successfully generalise from a labelled training dataset to predict the labels for unseen examples. 

%Motivating applications for classification include disease diagnosis, fraud detection, natural language processing and computer vision.

Notationally, each classification example has features ${\bf x}_i$ and a corresponding label $y_i$.
%Notationally, each classification example has a set of features ${\bf x}_i$ and a corresponding label $y_i$.
Thus each example is denoted by $ \{ {\bf x}_i, y_i\} $, where ${\bf x}_i$ is a $p$-dimensional feature vector, with a class label $y_i \in \{C_1, C_2, ..., C_k\}$.
%The true $k$ classes, also called \emph{labels}, are denoted by $C_i$ (missing for unlabelled examples).

%For a measles example, the true classes would be $\{C_1, C_2\}$, denoting presence/absence of the disease respectively;
%the features ${\bf x}_i$ would be patient symptoms: do they have red spots, coughing, high temperature, etc.

A dataset consists of $n$ examples, and is denoted $D = \{ {\bf x}_i$, $y_i\}^{n}_{1} $.
A classifier is an algorithm that predicts classes for unseen examples, with the objective of good generalisation on some performance measure.
A good overview of classification is provided by \cite[Chapter~1,2]{Tibshirani2009}.

\subsection{Active Learning}
The context for AL is where labelled examples are scarce or expensive.
For example in medical image diagnosis, it takes doctors' valuable time to label images with their correct diagnoses; but unlabelled
examples are plentiful and cheap. 
Given the high cost of obtaining a label, systematic selection of unlabelled examples for labelling might improve performance.
%The high cost of obtaining a label suggests that systematic selection of unlabelled examples for labelling might improve performance.
%The high cost of obtaining a label suggests that intelligent selection of unlabelled examples for labelling might improve performance.
An AL method can guide selection of the unlabelled data, to choose the most useful or informative examples for labelling.
In that way the AL method can choose unlabelled data to best improve the generalisation objective.
%If a heuristic can select the most useful or informative examples for labelling.
A small set of unlabelled examples is first chosen, then presented to an expert (\emph{oracle}) for labelling.
%In this AL scenario, a small set of unlabelled examples is first chosen, then presented to an expert (\emph{oracle}) for labelling.

Here we focus on batch AL; for variations, see \cite{Settles2009,Fu2013}.
A typical scenario would be a small number of initially labelled examples, a large pool of unlabelled examples, and a small budget of label requests.
An AL method spends the budget by choosing a small number of unlabelled examples, to receive labels from an oracle.
%Thus the work of any AL method is the selection of unlabelled examples from the pool.

An example AL method is uncertainty sampling using Shannon Entropy (denoted SE).
SE takes the entropy of the whole posterior probability vector for all classes. %, i.e. it applies the entropy function to the posterior probabilities.
Informally, SE expresses a distance metric of unlabelled points from the classifier decision boundary.

$$ 
Entropy = - \sum_{i=1}^{k} p(y_i | {\bf x}_j) \times \text{log} (p(y_i | {\bf x}_j)).
%Entropy = - \sum_{i=1}^{k} p(y_i | {\bf x}_j) \times log(p(y_i | {\bf x}_j)) = - [p(y_1 | {\bf x}_j) \times log(p(y_1 | {\bf x}_j)) + %p(y_2 | {\bf x}_j) \times log(p(y_2 | {\bf x}_j))]
%Entropy = - \sum_{i=1}^{k} p(y_i | {\bf x}_j) \times log(p(y_i | {\bf x}_j)).
$$

%An example AL method is uncertainty sampling using Shannon Entropy (denoted SE), described in Section %\ref{Results_from_a_second_AL_method_Marker}.
%Informally, SE expresses a distance metric of unlabelled points from the classifier decision boundary.
% (see \cite{Settles2009}).

Another example AL method is Query By Committee (denoted QBC), described in section \ref{Results_from_a_second_AL_method_Marker}. %random selection (denoted RS).
%Another example AL method is random selection of unlabelled examples (denoted RS).
%For an overview of AL contexts and methods, see \cite{Settles2009,Fu2013}. %REMOVED FOR SPACE

The oracle then satisfies those label requests, by providing the labels for that set of unlabelled examples.
%We consider a perfect oracle that immediately provides correct labels. %REMOVED FOR SPACE
The newly-labelled data is then combined with the initially-labelled data, to give a larger training dataset, to train an improved classifier.

The framework for AL described above is batch pool-based sampling; for variations see \cite{Fu2013,Settles2009}.
%There are many variations, for example AL with streaming classification (see \cite{Fu2013,Settles2009}).
%The simple framework for AL described above is batch pool-based sampling. 
%There are many variations, for example AL with streaming classification (see \cite{Fu2013,Settles2009}).

%
%\paragraph{Paragraph 1}
%Other sentence text here.
%

\subsection{Active Learning Factors}
\label{Active_Learning_Factors_Marker}

Intuitively there are several factors that might have an important effect on AL performance.
An experimental study can vary the values of those factors systematically to analyse their impact on AL performance.

% old todo revisit: need to MOTIVATE, why these factors matter and are interesting
One example of an AL factor is the nature of the classification task, including its difficulty and the complexity of the decision boundary.
The classifier can be expected to make a major difference, for example whether it can express linear and non-linear decision boundaries, and whether it is parametric.
The smoothness of the classification task input, for example continuous or discretised, might prove important since that smoothness affects the diversity of unlabelled examples in the pool.
Intuitively we might expect a discretised task to be harder than a continuous one, since that diversity of pool examples would decrease. 
%The smoothness of the classification task input, for example continuous or discretised, might prove important since that smoothness %affects the similarity spread of unlabelled examples in the pool.
Other relevant factors include the number of initial labels ($N_{initial}$) and the size of the label budget ($N_{budget}$). 

Some of these factors may be expected to materially determine AL performance.
%Intuitively we expect at least some of these factors to materially determine AL performance.
How the factors affect AL performance is an open question.
This experimental study evaluates AL methods for different combinations of factor values, i.e. at many points in factor space.
The goal here is to unravel how the factors affect AL performance.
%The goal here is to explore which factors most affect the performance of Active Learning.
A statistical analysis of the simulation study reveals some answers to that question, see Section \ref{Negative_Binomial_Regression_Analysis_Marker}.

Below the factor values are described in detail.

\begin{figure}
        \centering
        \begin{subfigure}[b]{0.2\textwidth}
                \centering
                \includegraphics[width=\textwidth]{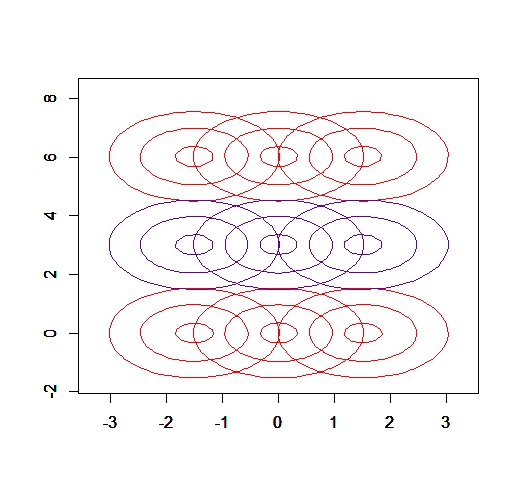}
                \caption{Task sd10}
                \label{fig3:Task sd10}
        \end{subfigure}
        \begin{subfigure}[b]{0.2\textwidth}
                \centering
                \includegraphics[width=\textwidth]{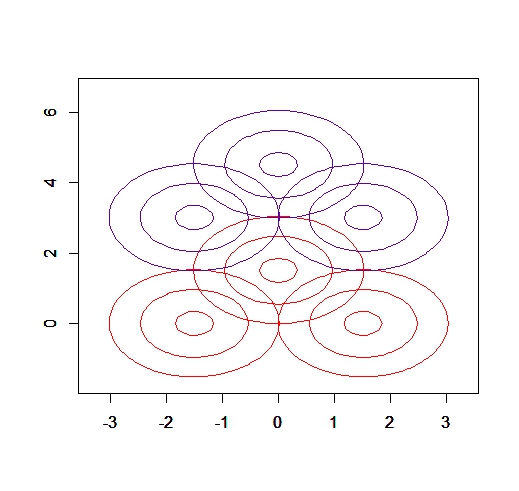}
                \caption{Task sd2}
                \label{fig3:Task sd2}
        \end{subfigure}%
        %~ %add desired spacing between images, e. g. ~, \quad, \qquad etc.
        \begin{subfigure}[b]{0.2\textwidth}
                \centering
                \includegraphics[width=\textwidth]{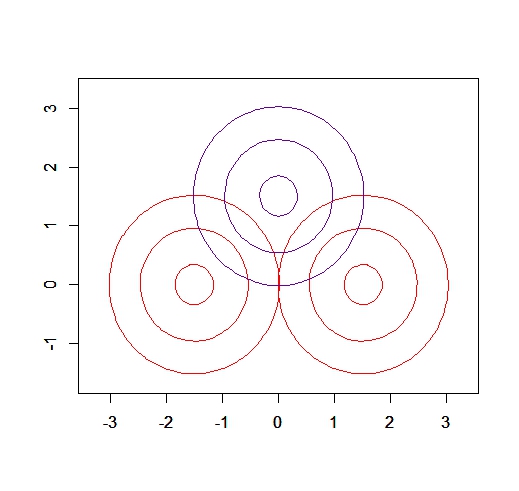}
                \caption{Task sd7}
                \label{fig3:Task sd7}
        \end{subfigure}
        \begin{subfigure}[b]{0.2\textwidth}
                \centering
                \includegraphics[width=\textwidth]{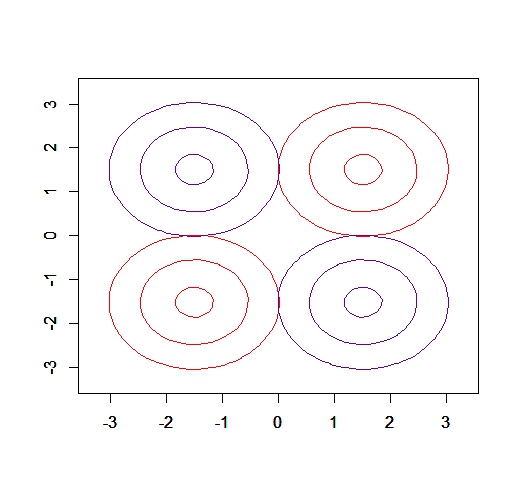}
                \caption{Task sd8}
                \label{fig3:Task sd8}
        \end{subfigure}
        \caption{Density contour plots to elucidate the classification problems}
	 %\caption{Classification Task contour graphs}
        %\caption{Contour graphs to show the tasks}
	  \label{fig3:Contour graphs to show the tasks}
\end{figure}

Four different simulated classification tasks are used, to vary the nature and complexity of the classification problem.
%The third task (sd7) is in many ways the simplest, with a nearly linear decision boundary.
%The fourth task (sd8) is a stochastic xor task, with a multiple-line boundary required to achieve the correct decision.
We restrict attention to binary classification problems.
%Figure \ref{fig3:Contour graphs to show the tasks} shows the tasks, which are formed from mixtures of Gaussian clusters.
Figure \ref{fig3:Contour graphs to show the tasks} shows the classification tasks.
%, which are formed from mixtures of Gaussian clusters.
These tasks are created from mixtures of Gaussian clusters.
The clusters are placed to create decision boundaries, some of which are simple curves and others are more involved.
In this way the complexity of the classification problem is varied across the tasks.

% very small graphs here to show the tasks.

Still focussing on the classification task, task difficulty is varied via the Bayes Error Rate (BER).
%Still focussing on the classification task, we also vary task difficulty via the Bayes Error Rate (BER). % as one way to quantify task difficulty.
Input smoothness is also varied, having the values continuous, discretised, or a mixture of both.
%Intuitively we might expect a discretised task to be harder than a continuous one. %moved
%, and hence explore the effect of that on AL performance.
BER is varied by modifying the Gaussian clusters for the problems; input smoothness is varied by transforming the realised datasets.

Another factor to vary is the input dimension $p$, by optionally adding extra dimensions independent of the class.
An interaction is expected between $p$ and the initial amount of labelled data $N_{initial}$, since higher dimensional data should require more datapoints to classify successfully.
%The input dimension is two. % old todo change

Four different classifiers were used: Logistic Regression (LogReg), Quadratic Discriminant Analysis (QDA), Random Forest (RF) and Support Vector Machines (SVM), to provide a variety of classifiers: linear and non-linear, parametric and non-parametric.
These classifiers are described in \cite{Tibshirani2009}.
The default parameters for RF are the defaults from R package RandomForest version 4.6-7; the default parameters for SVM are the defaults from R package RWeka version 0.4-14 (the complexity parameter C is chosen by cross-validation, the kernel is polynomial).

The amount of initial labelled data $N_{initial}$ is also varied. 
This factor is expected to be important, since too little data would give an AL method nothing to work with, and too much would often mean no possible scope for improvement.

The AL factors are summarised in Table \ref{table:AL Parameters Table 1}.
%The AL factors are summarised in the table below.
%This list was chosen as a first draft of the major relevant features.

%We have chosen a first set of AL factors that are summarised in the table below.
%A first set of AL factors is summarised in the table below.
%This list is chosen as a first draft of the major features that stand out as relevant.

%Note, used to show one example, now show whole set of values
\begin{table}[h!b!p!] %[ht]
\caption{Active Learning Factors}
\label{table:AL Parameters Table 1}
\centering
\begin{tabular}{|c|c|}
\hline
Name & Values\\
\hline
\hline
\makecell{Classification Task} & sd10, sd2, sd7, sd8 (see Figure \ref{fig3:Contour graphs to show the tasks})\\
%\makecell{Classification Task} & Tasks described above\\
\hline
\makecell{Task Input Type} & Continuous, Discretised, Mixed\\
\hline
\makecell{Task Input Dimension} & 2, 10\\
\hline
\makecell{Classifier} & LogReg, QDA, RF, SVM\\
\hline
\makecell{ $N_{initial}$ } & 10, 25, 50, 100\\
%\makecell{Number of initially labelled examples, $N_{initial}$ } & Numeric (non-negative integer) & 10, 25, 50, 100\\
\hline
\makecell{Bayes Error Rate} & 0.1, 0.2, 0.35\\
\hline
\makecell{Classifier Optimum Error Rate} & [inferred]\\
\hline
\makecell{Space for AL} & [inferred]\\
\hline
%\makecell{Multilined \\ cell text} & 28--31\\
%\hline
\end{tabular}
\end{table}

% Note - random seed is varied to ensure repeatability, but not mentioned since not interesting
% Note - alChunkSize is fixed at 1 for prob tasks, so does not vary; note, this varies for data task results
% Note - input dimension is fixed at 2 for prob tasks, so does not vary
% Note - num classes is fixed at 2 for prob tasks, so does not vary

%REMOVED FOR SPACE
%Two of these factors, classifier optimum error rate and space for AL (defined below), are inferred quantities rather than factors whose %values are chosen in advance. 

%, see Section \ref{Aggregating_results_across_Experiments_Marker}.
%For a particular experiment we also infer features that are not factors, for example classifier-task optimum error rate and the space %for AL, see Section \ref{Aggregating_results_across_Experiments_Marker}.

The optimum error rate for the classifier on a specific task is evaluated experimentally, by averaging the results of several large train-test datasets, to provide a ceiling benchmark.
%This optimum provides a ceiling benchmark, the best that the classifier can do for this task.
%%Some other experimental attributes are then calculated or inferred.
%The optimum error rate for the classifier on a specific task is evaluated experimentally, by averaging the results of several large train-test %datasets.
%This optimum provides a ceiling benchmark, the best that the classifier can do for this task.
%%This optimum provides a ceiling benchmark, the best that the classifier can do for this task; this optimum is bounded above by BER.

We also consider the potential space for AL to provide a performance gain.
In the context of simulated data all labels are known, and some labels are hidden to perform the AL experimental study.
The classifier that sees all the labelled data provides a ceiling benchmark, the score $ S_{all} $.
The classifier that sees only the initially labelled data provides a floor benchmark, the score $ S_{initial} $.
To quantify the scope for AL to improve performance, we define the space for AL as a ratio of performance scores: $ (S_{all} - S_{initial}) / S_{all} $.
This provides a normalised metric of the potential for AL to improve performance.

A Monte Carlo experiment varying these factors provides the opportunity to statistically analyse the behaviour of AL.
To get to this point, both a careful experiment and a refined methodology of performance assessment are required.

\section{Experimental Method}
\label{Experimental_Method_Marker}

%Here we describe the experimental setup.

AL is applied iteratively in these experiments:  the amount of labelled data grows progressively, as the AL method spends a budget chunk at each time point.
We may choose to spend our overall budget all in one go, or iteratively, in smaller batches. 
%Thus the amount of labelled data grows incrementally; at each time point, a new budget chunk is spent on label requests.

%For an AL experiment, AL is applied iteratively: the amount of labelled data grows progressively, as the AL method progressively spends %budget chunks to increase the amount of labelled data.
%$ D_{labelled}$, up to a maximum budget size $N_{MaxBudget}$.
%Thus the amount of labelled data grows incrementally; at each time point, a new budget is spent on label requests.

In that sense the experimental setup resembles that of the AL challenge described in \cite{Guyon2011}.
We use this iteration over budget because it is realistic for practical AL applications, and because it explores the behaviour of AL as the number of labelled examples grows.
%Further, this iteration allows a direct comparison between the empirical learning curves of an AL method and random selection (RS).
Experiments consider AL methods SE and QBC.
%The experiments examine the popular AL method uncertainty sampling using Shannon Entropy (SE), chosen for its simplicity and %popularity.
%A more theoretically motivated AL method, QBC, is also evaluated experimentally (see Section %\ref{Results_from_a_second_AL_method_Marker}).
%The performance metric used is success rate.

\begin{figure}
        \centering
	 \begin{subfigure}[b]{0.52\textwidth}
        %\begin{subfigure}[b]{0.7\textwidth}
        %\begin{subfigure}[b]{0.56\textwidth}
                \centering
                \includegraphics[width=\textwidth]{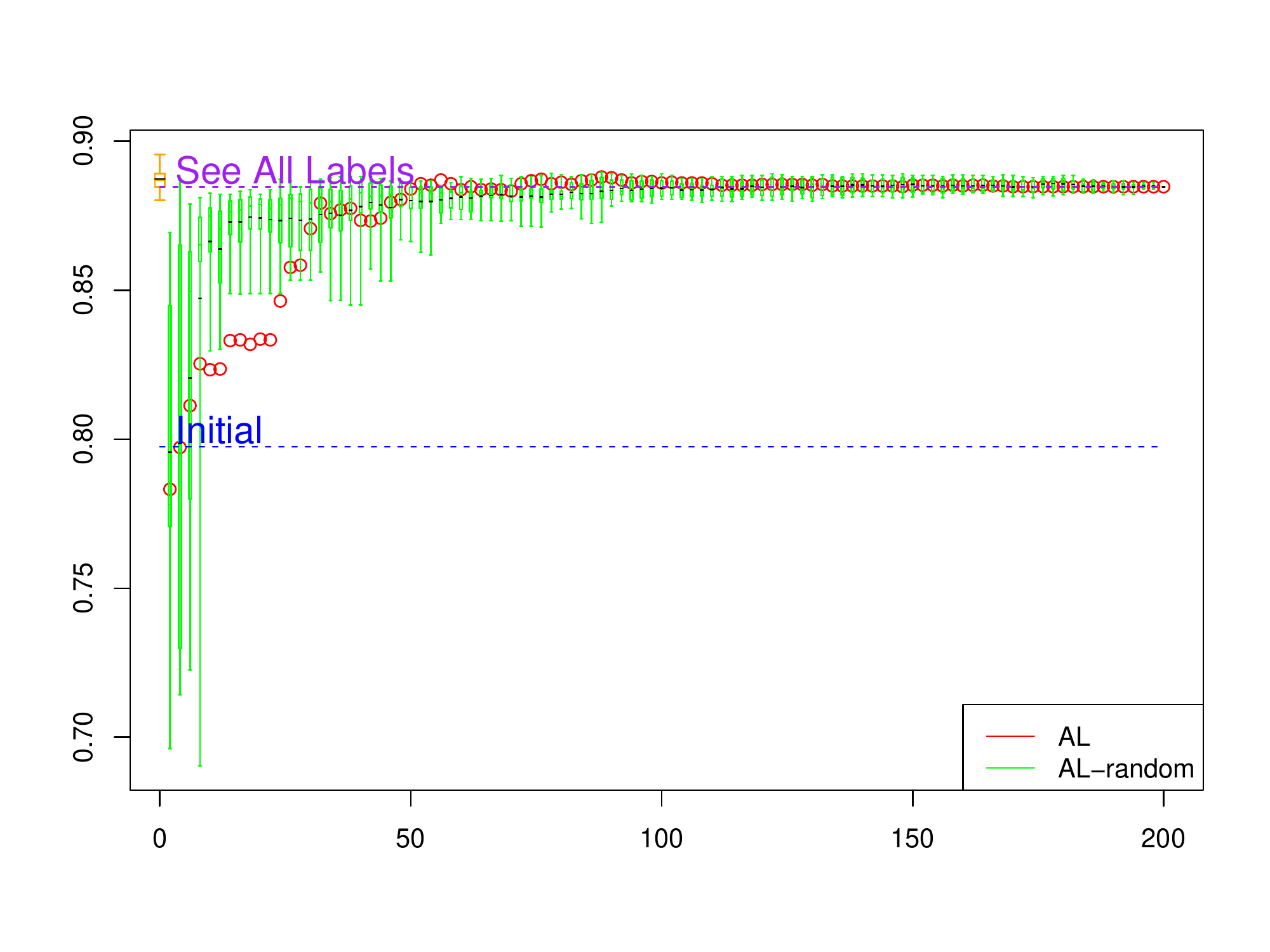}
                \caption{Scores $ \boldsymbol{S_i} $}
                \label{fig1:Scores}
        \end{subfigure}%
        %~ %add desired spacing between images, e. g. ~, \quad, \qquad etc.
        \begin{subfigure}[b]{0.52\textwidth}
	 %\begin{subfigure}[b]{0.7\textwidth}
        %\begin{subfigure}[b]{0.56\textwidth}
                \centering
                \includegraphics[width=\textwidth]{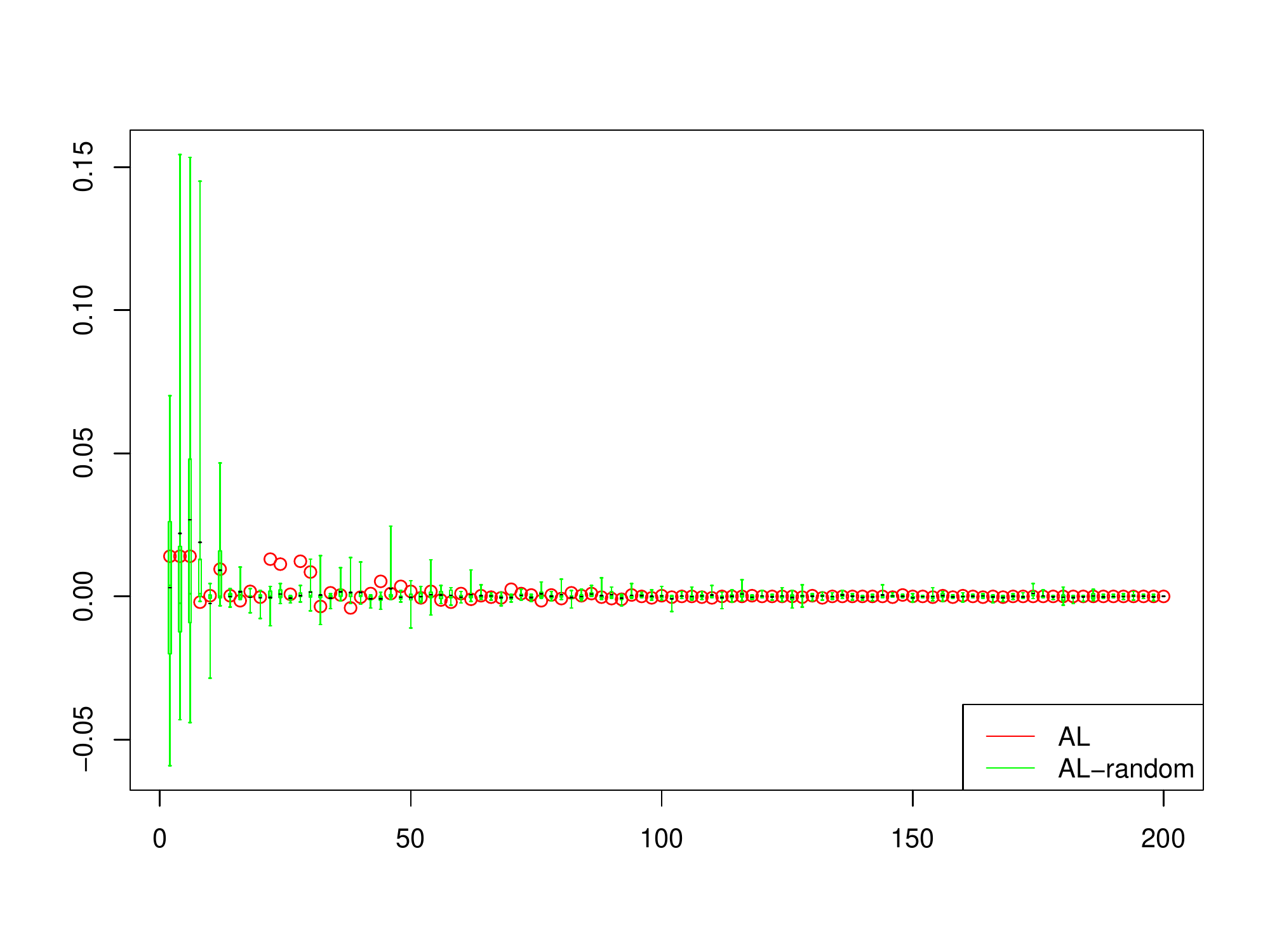}
		  \caption{Score differences $ \boldsymbol{\delta S_i} $}
                \label{fig1:Score deltas}
        \end{subfigure}
        \caption{Scores $ \boldsymbol{S_i} $ in subgraph (a) and score differences $ \boldsymbol{\delta S_i} $ in subgraph (b), for AL method Shannon Entropy vs Random Selection}
	\label{fig1:Scores and score differences for AL method vs Random Selection}
\end{figure}

To motivate our experimental method, we present the summary plots of the relative performances of AL and RS over time, see Figures \ref{fig1:Scores} and \ref{fig1:Score deltas}.

% possible ref here for "Modelling Classification Performance for Large Data Sets - An Empirical Study___10.1.1.28.8876.pdf"

The experimental setup is as follows.
%The experimental setup is expressed algorithmically.
Firstly, sample a pair of datasets [$D_{train}$, $D_{test}$] from the classification task.
%Firstly, sample a pair of datasets $[D_{train}, D_{test}]$ from the classification task.
To simulate label scarcity, split the training dataset into initally labelled data $ D_{initial}$ and an unlabelled pool $ D_{pool}$.
%Then for label scarcity, split the training dataset into initally labelled data $ D_{initial}$ and an unlabelled pool $ D_{pool}$.

% REMOVED as repetitive, and for space
%The experiment involves an iteration over many budget values. 
%The budget progressively grows, and AL performance is evaluated at each successive budget point. % (time point). 

The output of one experiment can be described in a single plot, for example Figure \ref{fig1:Scores}.
That figure shows the trajectory of performance scores obtained from progressive labelling, as follows.
At each time point the AL method chooses a small set of examples for labelling, which is added to the existing dataset of labelled data.
This selection happens repeatedly, creating a trajectory of selected datasets from the unlabelled pool.
Each time point gives a performance score, for example error rate, though the framework extends to any performance metric.
This gives the overall result of a trajectory of scores over time, denoted $ \boldsymbol{S_i} $: an empirical learning curve.
Here $i$ denotes the time point as we iterately increase the amount of labelled data, with $i \in [0,100]$.

Given several instances of RS, we form an empirical boxplot, called a sampling interval.
Figure \ref{fig1:Scores} shows the trajectory of scores for the AL method, and the vertical boxplots show the sampling intervals for the scores for RS.
%The figure shows both the trajectory of scores for the AL method, and the vertical boxplots show the sampling intervals for the scores for RS.

Once this iterative process is done, we obtain a set of scores over the whole budget range, denoted $ \boldsymbol{S_i} $. 
%Once the iteration of $N_{budget}$ is done, we have the scores over the whole budget range, denoted $ \boldsymbol{S_i} $.
Those scores are used to calculate various performance comparisons, specifically to see whether AL outperformed RS, see Section \ref{Methodology_Marker}.

The AL method now has a score trajectory $ \boldsymbol{S_i} $: a set of scores over the whole budget range.
All trajectories begin at the floor benchmark score $S_{initial}$ and terminate at the ceiling benchmark score $S_{all}$. 
From the score trajectory $ \boldsymbol{S_i} $ a set of score differences $ \boldsymbol{\delta S_i} $ is calculated via $ \boldsymbol{\delta S_i} = \boldsymbol{S_i} - \boldsymbol{S_{i-1}}$.
The need for and usage of the score differences is detailed in Section \ref{Methodology_Marker}.
The chosen AL method is evaluated alongside several instances of RS, the latter providing a benchmark.
Experiments are repeated to generate several instances of RS, since RS shows substantial variability.

To illustrate the trajectories of the performance scores $ \boldsymbol{S_i} $, Figure \ref{fig1:Scores} shows those scores for the AL method SE and comparison with RS. %the instances of RS.
%The score differences are shown in Figure \ref{fig1:Score deltas}.
%%Figures \ref{fig1:Scores} and \ref{fig1:Score deltas} show the raw performance scores and the score differences for the AL method 
%%and the instances of RS.

% old score syntax
% $ ({S_i}) $
% $ ({S_i})^n_1 $
\subsection{Methodology to evaluate AL performance}
\label{Methodology_Marker}

%There are several important methodological subtleties with the assessment of AL performance.
%These subtleties include the high varability of the performance scores for both AL methods and random selection, and the relative weakness of AL performance gain for many classification problems.
%
%This section elucidates the difficulties with existing AL performance metrics, and contributes a novel assessment metholodology to address those complexities.
%%This methodology addresses the questions of where AL works and how much.
%The primary goal of every AL performance metric is to quantify the AL performance gain from a given experiment, as a single scalar summary.
%%That performance gain is calibrated against random selection.

This section elucidates the difficulties with existing AL performance metrics, and contributes a novel assessment metholodology to address those complexities.
The primary goal of every AL performance metric is to quantify the AL performance gain from a given experiment, as a single scalar summary.

Any AL performance methodology must first address two preliminary issues: the benchmark for AL to outperform, and how to handle the variability of that benchmark.
The first issue is to decide which benchmark should be used to compare AL methods against.
One option is to compare AL performance to the initial classifier.
However, that ignores the fact that the labelled dataset is larger in the case of AL: even random selection of further examples for labelling would be expected to improve performance on average, since the classifier sees a larger training dataset. 
%Thus the correct benchmark for AL is random selection (RS), that sees exactly the same amount of labelled data as the AL method.
Thus a better benchmark for AL is random selection (RS), that sees exactly the same amount of labelled data as the AL method.

The second issue concerns the high variability of the benchmark, given that experiments show RS to have high variability.
The approach used here is to evaluate multiple instances of RS, to get a reasonable estimate of both location and dispersion of performance score.
From those multiple instances we can form a sampling interval of the RS score, and thus capture its high variability.

\subsubsection{Score trajectories under experimental budget iteration}

Having established the benchmark of RS, we consider the score trajectories in the experimental context of budget iteration, to better understand how to compare AL against its benchmark.

We begin with the score trajectories $ \boldsymbol{S_i} $ derived from the budget iteration process.
The budget is iterated over the entire pool in 100 steps; during that iteration, the amount of labelled data grows from its minimum $N_{initial}$ to its maximum $N_{train}$.
At each budget iteration step, the available budget is small compared to the total size of the pool.
This is illustrated in Figure \ref{fig1:Scores}.

Each score trajectory $ \boldsymbol{S_i} $ has significant autocorrelation, since each value depends largely on the previous one.
To see this for the score trajectory, recall that for each budget iteration step, the available budget is small. 
%Thus the score at a given step is mainly determined by the score at the previous step.
Hence the score at one step $ S_i $ is very close to the score at the previous step $ S_{i-1} $.
Thus the scores $ \boldsymbol{S_i} $ only change incrementally with each budget iteration step, giving rise to high degrees of autocorrelation. 

In contrast, the score differences $ \boldsymbol{\delta S_i} $ are expected to be substantially less autocorrelated.
%In contrast, the score differences $ \boldsymbol{\delta S_i} = (S_i) - (S_{i-1})$ are expected to be substantially less autocorrelated.
This belief is confirmed experimentally by ACF graphs, which show significant autocorrelation for the scores but not for the score differences.
This contrast matters when comparing different AL performance metrics.
%make a big difference to the comparison of different AL performance metrics below.
%This point about autocorrelations matters for the comparison of different AL performance metrics below.

%One analogy is with a stochastic process or time series, where the increments are independent but the values are not.
%Thus the values are autocorrelated whereas the increments are not. 
%%%, allowing the increments an independence that is denied to the values.

%To see this for the score trajectory, recall that for each budget iteration step, the available budget is small. 
%%%Thus the score at a given step is mainly determined by the score at the previous step.
%Hence the score at one step $ [ S_i ] $ is very close to the score at the previous step $ [ S_{i-1} ] $.
%Thus the scores $ \boldsymbol{S_i} $ only change incrementally with each budget iteration step, giving rise to high degrees of autocorrelation. 
%%%Thus the scores $ \boldsymbol{S_i} $ only change incrementally with each new AL dataset selection, giving rise to high degrees of autocorrelation. 

\subsubsection{Comparing AL performance metrics}

We now address different AL performance metrics, each designed to measure the performance of AL methods.
Two common AL performance metrics are direct comparisons of the score trajectories, and the Area Under the Active learning curve (AUA) (see \cite{Guyon2011}).

The autocorrelation of score trajectories $ \boldsymbol{S_i} $ means that directly comparing two score trajectories is potentially misleading.
For example, if an AL method does well against RS only for a small time at the start, and then does equally well, this would lead to the AL method's score trajectory dominating that of the RS over the whole budget range.
This would present a false picture of where the AL method is outperforming RS.
Much of the AL literature suggests that this early AL performance zone is precisely to be expected (\cite{Settles2009}), and thus this comparison may often be partially flawed.
Further, this same case shows that the AUA (see \cite{Guyon2011}) would overstate the AL performance gain; see Figure \ref{fig1:Scores} which shows the score trajectories.

%$\textbf{aaa}$
%\textbf{$\delta S_i$}
Here we resolve that problem by considering the score differences $ \boldsymbol{\delta S_i} $, not the scores themselves $ \boldsymbol {S_i} $.
%The approach taken here to resolve that problem is.....
%One way in this study to resolve that issue is to consider the score differences $ \boldsymbol{\delta S_i} $, not the scores themselves $ \boldsymbol{S_i} $.
Those differences show much less autocorrelation than the scores (this is shown by ACF graphs).

An example of the score differences $ \boldsymbol{\delta S_i} $ is shown in Figure \ref{fig1:Score deltas}.
Our new methodology is based on examining these score differences.
%Hence our new methodology is based on examining these score differences.
%, given the reasons above for potential pitfalls with the AL performance methods of directly comparing scores, and AUA}.
%Given the reasons above for rejecting the AL performance metrics of directly comparing scores, and AUA, our methodology is %based on examining these score differences.

\subsubsection{A new methodology to evaluate AL performance}

Our new methodology is based on comparing the score differences $ \boldsymbol{\delta S_i}  $, as a way to compare AL against its benchmark RS.
This is done in two stages.

The first stage is to seek a function that quantifies the result of the comparison between two score differences, $ {\delta S_i}^{SE} $ for AL method SE and $ {\delta S_i}^{RS} $ for RS.
To ensure fair comparisons, ties need to be scored differently to both wins and losses.
The approach adopted here is to use a simple comparison function $f$: 

\begin{displaymath}
   f(x,y) = \left\{
     \begin{array}{lr}
       1 & : x > y\\
       0.5 & : x = y\\
       0 & : x < y.\\
     \end{array}
   \right.
\end{displaymath} 

This comparison function $f$ is applied to two score differences, e.g. $ f ( {\delta S_i}^{SE} ,  {\delta S_i}^{RS} ) $.
The motivation here is to carefully distinguish wins, losses and ties, and to capture those three outcomes in one scalar summary.
Applying that comparison function to compare all the score differences of SE and RS generates a set of comparison values, denoted $ \boldsymbol{C_i} $, each value $ \in [0,1] $.
Several instances of RS generate several such sets of values, one for each instance. % i.e. $ (C_i) , i \in [1,N_{RS}]$.

We use several instances of RS to capture its high variability, the number of RS instances being $ N_{RS} $.
Each instance $j$ has its own set of comparison scores $ \boldsymbol{C_i^j} $. %, j \in [1, N_{RS}] $.
%Each instance has its own set of comparison scores $ \boldsymbol{C_i} $.
%%%%%We use several instances of RS to capture its high variability; thus we need to average the results from comparing AL to each %instance.
Those comparison values $ \boldsymbol{C_i^j} $ are then averaged to form a single set of averaged comparison values, denoted $ \boldsymbol{A_i} =
\frac{1}{N_{RS}} \sum_{j=1}^{N_{RS}} \boldsymbol{C_i^j}$. 
Further, each value $ \boldsymbol{A_i} \in [0,1] $.
%1/N_{RS} \sum_{j=1}^{N_{RS}} C_i^j$.
%Those comparison values $ \boldsymbol{C_i} $ are then averaged to form a single set of averaged comparison values, denoted $ \boldsymbol{A_i} $, %each in [0,1].

\begin{figure}
        \centering
	 \includegraphics[width=0.75\textwidth]{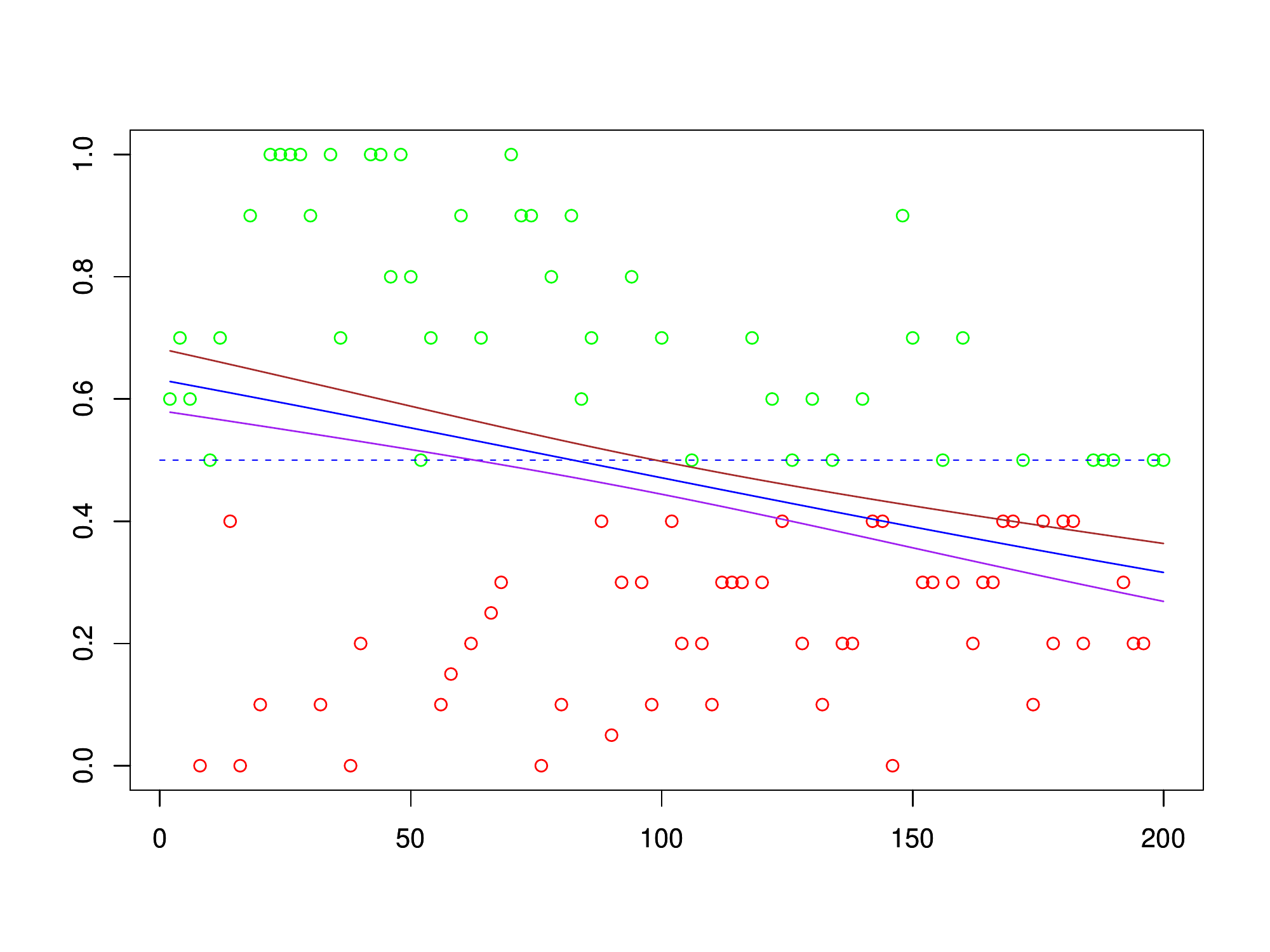}
        \caption{Averaged Comparison Values $ \boldsymbol{A_i} $ with Generalised Additive Model curve and pointwise 80\% confidence interval}
        %\caption{Averaged Comparison Values $ \boldsymbol{A_i} $ with Generalised Additive Model}
	\label{fig2:Averaged Comparison Scores with GAM}
\end{figure}

%%%The score differences are then averaged to form a single set of averaged comparison values, denoted $ \boldsymbol{A_i} $, each in [0,1].
That single set of values $ \boldsymbol{A_i} $ provides a summary of the overall performance comparison between the AL method and RS.
%That single set of values $ \boldsymbol{A_i} $ can then summarise the overall comparison between the AL method and RS.
%%%Since we have several instances of RS, we average the comparison values over the $n$ RS.
%%%That gives a mean comparison value in [0,1], to indicate how well the AL method did against $n$ instances of RS.
%%%This results in a mean comparison value for every time point, i.e. for each value of $N_{budget}$.
That comparison is illustrated in Figure \ref{fig2:Averaged Comparison Scores with GAM} which shows those average comparison values  $ \boldsymbol{A_i} $ over the whole budget range.
%Figure \ref{fig2:Averaged Comparison Scores with GAM} illustrates those average comparison values  $ \boldsymbol{A_i} $ over the whole %budget range.

The final stage of the new method is interpreting the averaged comparison values $ \boldsymbol{A_i} $. %, each in [0,1].
The aim is to extract the relationship between $ \boldsymbol{A_i} $ and budget, with a confidence interval band.
%The aim here is to extract the true curve, but also with a slightly pessimistic interpretation, to avoid misinterpreting chance effects as AL %performance gains.
%Thus the aim is to extract the true curve, but also with a confidence interval band.

The lower 80\% confidence interval is chosen to form a mildly pessimistic estimate of AL performance gain.
We fitted a Generalised Additive Model (GAM) to this set of values (given the need for inference of confidence intervals).
The GAM is chosen using a logit link function, with variable dispersion to get better confidence intervals under potential model mis-specification (see \cite{Hastie1990}).
The GAM is implemented by R package mgcv version 1.7-22; the smoother function default is thin plate regression splines.
The GAM relates the expected value of the distribution to the covariates thus:

%%%%The goal is to extract the true curve, but also with a confidence interval band, 
%%%%to form a conservative estimate of the location of the true curve.
%Given this need for inference, one reasonable choice is to fit a Generalised Additive Model (GAM) to this set of values.
%The lower 80\% confidence interval is chosen to form a mildly pessimistic estimate of AL performance gain.
%The GAM is chosen using a logit link function, with variable dispersion to get better confidence intervals under potential model mis-specification (see \cite{Hastie1990}).
%%%%The GAM chosen is Logistic Regression, with variable dispersion to get better confidence intervals under potential model mis-specification %%%(see \cite{Hastie1990}).
%The default GAM from R library mgcv version 1.7-22; the smoother function default is thin plate regression splines.
%%%%The GAM chosen is quasi-binomial; the link function is logit; the scale of the response is in [0,1]; the datapoints are equally weighted.

\begin{equation*}
g(\operatorname{E}(Y))=\beta_0 + \beta_1 f_1(x_1).
%g(\operatorname{E}(Y))=\beta_0 + \beta_0 \times f_1(x_1)
%g(\operatorname{E}(Y))=\beta_0 + f_1(x_1) + f_2(x_2)+ \cdots + f_m(x_m)
\end{equation*}

%We fit a Generalised Additive Model (GAM) to this set of values (see \cite{Hastie1990}), and take the lower 80\% confidence interval curve %as a conservative estimate of the location of the true curve. 
The fitted GAM is shown in Figure \ref{fig2:Averaged Comparison Scores with GAM}.
The estimated effect seems roughly linear.
The baseline level of 0.5 is shown as a dotted line, which represents an AL method that ties with RS, i.e. does not outperform it.

Given the intricacies of evaluating AL performance, a primary goal for this methodology is to quantify AL performance from a given experiment as a single result.
The GAM curve shows where the AL performance zone is, namely the initial region where the curve is significantly above 0.5.
We consider the initial region, as much AL literature suggests that the AL performance gain occurs early in the learning curve, see \cite{Settles2009}.
Thus the length of the AL performance zone is the single result that summarises each experiment.

Overall, this methodology addresses some of the complexities of assessment of AL performance in simulation contexts.
As such it provides a milestone on the road to more accurate and statistically significant measurements of AL performance.
This is important given that many authors find that the AL performance effect can be elusive (e.g. \cite {Guyon2011,Cawley2011}).
%This methodology could also be used retrospectively to confirm the results of existing AL studies and surveys.

This methodology is illustrated with specific results in Figures  \ref{fig1:Scores}, \ref{fig1:Score deltas} and \ref{fig2:Averaged Comparison Scores with GAM}.
%This methodology is illustrated with specific results below.
%Figures \ref{fig1:Scores} and \ref{fig1:Score deltas} show the performance scores $ \boldsymbol{S_i} $ and the score differences $ \boldsymbol{\delta S_i} $ for the AL method and the instances of RS.
%Figure \ref{fig2:Averaged Comparison Scores with GAM} show the averaged comparison values $ \boldsymbol{A_i} $ with the fitted GAM, for %the AL method and the instances of RS.

%Figures \ref{fig2:Averaged Comparison Scores} and \ref{fig2:Averaged Comparison Scores with GAM} show the averaged comparison %scores, and those scores with the fitted GAM, for the AL method and the instances of AL-Random.

%%%The results for all experiments are aggregated together into one overall dataset.
%%%The form of that overall dataset is a regression dataset: several features and a single numerical response.
%The form of the aggregate result dataset is a regression dataset: several factors and a single numerical response. % REMOVED
%%%That dataset is then analysed, to explain the variability in AL performance as a function of the factors.
%%%%See Section \ref{Results_Marker} below.

\section{Results and Discussion}
\label{Results_Marker}

%The aggregate result dataset has the form of a regression dataset.
The dependent variable is the AL performance zone length, an integer count.
That value is obtained via the methodology described above, which includes fitting a GAM to ensure statistical significance.
The factors are given in Table \ref{table:AL Parameters Table 1}.
%We analyse the aggregate result dataset, which has the form of a regression dataset, using a GLM model.

\subsection{Negative Binomial Regression Analysis}
\label{Negative_Binomial_Regression_Analysis_Marker}

% The primary analysis of the aggregate result dataset is negative binomial regression.
The experimental output includes the AL performance zone length (derived from the GAM) to the AL factors.
Given the form of the aggregate experimental output, the appropriate initial analyses were Poisson and Negative-Binomial regression. 
A Poisson regression model was found to be over-dispersed.
%We fit a negative binomial regression generalised linear model, which fits reasonably well with greatly improved fit, with a %deviance/df.residual ratio of 0.327 and dispersion of 0.142.
We fit a negative binomial regression generalised linear model, which fits reasonably well with significant under-dispersion:
%We fit a negative binomial regression model, which fits reasonably well with a deviance/df.residual ratio of 0.327 and dispersion of 0.142.

% neg-bin-regression equation here
\begin{equation*}
Y_i \sim \textup{NegBin}(\mu_i, \kappa)
\end{equation*}
with
\begin{equation*}
log(\mu_i) = {\bf x}_i \cdot {\boldsymbol{\beta}}
%log(\mu_i) = x_i \times \beta
%log(\mu_i) = x_i * \beta
\end{equation*}

where $\kappa$ is a dispersion parameter.

The significant results of that model are summarised in Table \ref{table:Negative Binomial Result 1}.

\begin{table}[h!b!p!]
\caption{Negative Binomial Significant Results}
\label{table:Negative Binomial Result 1}
\centering
%\begin{tabular}{|c|c|c|}
\begin{tabular}{| c | . | c |}

%\begin{tabular}{@{}l*{2}{D{.}{.}{7}}@{}}

%\begin{tabular}{|c| >{\centering\arraybackslash}m{1in} |c|}
%\begin{tabular}{S[table-format=3.2]}{|c|c|c|}
%\begin{tabular}{S[table-format=3.2]}
%\begin{tabular}{@{}l S[table-format=3.2] S[table-format=3.2]@{}}
%\begin{tabular}{@{}l S[table-format=3.2] S[table-format=3.2]@{}}
\hline
Name & \textup{Coefficient} & p-value\\
%\textbf{Name} & \textbf{Coefficient} & \textbf{p-value}\\
%Name & Coefficient & p-value\\
\hline
\hline
\makecell{Intercept} & -1.695 & 1.70e-12\\
\hline
\makecell{Specific Classifier, Logistic Regression} & 1.142 & \textless 2e-16\\
\hline
\makecell{Specific Task, labelled sd7} & -0.481 & 0.000405\\
\hline
\makecell{Input Type, Continuous} & 0.578 & 1.11e-09\\
\hline
\makecell{Input Type, Discrete} & -1.235 & \textless 2e-16\\
\hline
%\makecell{Bayes Error Rate} & -1.845 & 0.0423\\
%\hline
%\makecell{Classifier Optimum Error Rate for Task} & 1.853 & 0.0327\\
%\hline
%\makecell{Multilined \\ cell text} & 28--31\\
%\hline
\end{tabular}
\end{table}

There are several results from the negative binomial regression which were not anticipated.
%For example, classifier proved significant: Logistic Regression gives significantly more AL performance than SVM.
%For example, Logistic Regression gives significantly more AL performance than SVM.
For example, LogReg shows more improvement, all things being equal, than SVM, for AL method SE.
%Why should this be true, given that SVM is widely regarded as the more flexible and powerful of the two classifiers?

This may be due to classifier mis-match: one might conjecture that AL works better when the classifier is mis-matched to the task, because the range of example quality within the unlabelled pool might be much higher under mis-match. % than under good specification.

Here classifier mis-match means the experimental metric of the classifier's sub-optimality on a given task.
Classifier mis-match is the performance difference between this classifier and the optimal Bayes classifier on the task.
Informally,  mis-match measures how ill-suited a classifier is to a given task.
%Informally we can use mis-match as a measure of how ill-suited the classifier is to the given task.

Under correct classifier match, most examples will improve a classifier's performance, whereas under mis-match, some examples may reduce the performance while others improve it, leading to a greater range of example quality under mis-match.

%%One answer might lie with classifier mis-specification: one theory is that AL works better when the classifier is mis-specified, as the range %of %example quality for examples within the unlabelled pool is much higher under mis-specification than under good specification.
%% old todo need a ref here.
%% had added this, but then realised, have already said this, in paragraph below
%%That theory is consistent with the result that Logistic Regression is significantly better for AL benefit than SVM.
%%It is also consistent with the result that the simplest task (labelled sd7) gave significantly less benefit to AL.

%To probe this theory further, future work will examine specific combinations of classifier and task that vary the degree of mis-specification %systematically, and examine the range of example quality in the pool.

%The tasks are described in Section \ref{Active_Learning_Factors_Marker}.
%The third task (sd7) is in many ways the simplest, with a nearly linear decision boundary.
%The fourth task (sd8) is a stochastic xor task, with a multiple-line boundary required to achieve the correct decision.

The choice of task is significant: the third task is worse than the fourth.
The fourth task has a more complex decision boundary than the third, leading to expected greater model mis-match for this task.
%The fourth task has a more complex decision boundary than the third, leading to expected greater model mis-specification for this task. Section \ref{Active_Learning_Factors_Marker} describes the tasks.
The fact that the third is worse for AL than the fourth is also consistent with the conjecture described above, that AL works better under mis-match.
%The fact that the third is worse for AL than the fourth is also consistent with the conjecture described above, that AL works better under %mis-specification.
%The fact that the third is worse for AL than the fourth is also consistent with the mis-specification theory of AL.
%The fact that the third is worse for AL than the fourth is also consistent with the mis-specification theory of AL.

%The choice of task is significant: the third task is worse than the fourth (Section \ref{Active_Learning_Factors_Marker} describes the tasks).
%The fact that the third is worse for AL than the fourth is also consistent with the mis-specification theory of AL.

% REMOVED for new results with p2 and 10
%Classifier optimum error rate is found to be significant, with a positive effect on AL.
%The larger this value, the more performance AL can provide, all other things being equal.
%There is a theory that AL only works with correctly specified classifiers (\cite{Cohn1996}); this experimental result confirms that theory.

%$N_{initial}$ and spaceForAL both turn out to not be significant factors.

%There is a widespread belief in the AL literature that the AL performance zone is early on in the budget range.
There is a widespread belief in the AL literature that the AL performance zone is early in the budget range (see \cite{Settles2009}).
In other words, as we progressively increase the amount of the labelled data, AL provides its performance gain earlier more than later.
%We expect AL methods to select the more useful examples from the pool, and the greatest range of usefulness would exist early on.
AL methods are expected to select the more useful examples from the pool, and the greatest range of usefulness would exist early on.
In practical applications, AL is usually required work earlier rather than later, since the essential context of AL is label scarcity.
%In practical applications, we usually require AL to work earlier rather than later, since the whole context of AL is label scarcity.
This belief is confirmed by the analysis: for the experiments that showed an AL performance gain, the mean and median lengths of the AL performance zone length were 38 and 32 respectively, out of a maximum of 200. 
%These al zone length values are slightly positively skewed (the skewness value is 0.63).
%This belief is confirmed by this analysis: for the experiments that showed an AL performance gain, the mean and median lengths of the %AL performance zone length were 34 and 39 respectively, out of a maximum of 200. 
%These al zone length values are slightly positively skewed (the skewness value is 0.52).

It is notable that input dimension turns out not to be significant.

It was quite rare for AL to show a performance gain at all, compared to RS, only in around 11\% of experiments.
%How often did AL show a performance gain at all, compared to random selection? 
%Quite rarely: only in around 11\% of experiments.
This confirms existing studies that the AL performance gain is often elusive (\cite{Baldridge2004,Cawley2011,Provost2010}).
It also emphasises the clear need for a precise reasoned methodology to analyse AL performance, hence the detailed methodology described in Section \ref{Methodology_Marker}.

\subsection{Results from QBC}
\label{Results_from_a_second_AL_method_Marker}

%The results described above were generated using the AL method Shannon Entropy (SE), a form of uncertainty sampling. 
% (\cite{Settles2009}).
To explore the importance of the AL method used, experiments evaluated a different AL method, QBC, using average KL-divergence as the disagreement measure. % to evaluate the combined predicted class-membership probabilities.
The two AL methods SE and QBC are very different in both algorithmic details and overall motivation (see \cite{Settles2009,Fu2013}), making it worthwhile to compare their results.
%Overall, QBC showed a benefit quite rarely: only in around 6\% of experiments.

The AL method QBC takes a committee of classifiers, and scores an unlabelled example ${\bf x}_j$ by how much disagreement there is within the committee.
Disagreement measures include Vote Entropy and Average K-L Divergence; see \cite{Fu2013,Settles2009}.
For QBC the classifier committee was Logistic Regression, k-nearest-neighbour (with $k=5$ and $k=21$), Support Vector Machine and Random Forest.
%For QBC the classifier committee was Logistic Regression, Knn (k=5), Knn (k=21), Support Vector Machine and Random Forest.

The experimental setup was identical, and the results were analysed in the same way: by a negative binomial regression analysis.
That model fits reasonably well. %, with very good fit. %a deviance/df.residual ratio of 0.171 and dispersion of 0.043.
%That model fits reasonably well, with significant under-dispersion. %a deviance/df.residual ratio of 0.171 and dispersion of 0.043.
The results from QBC are somewhat different to those from the SE.
%The significant results from that analysis of QBC are somewhat different to those from the SE in Section \ref{Negative_Binomial_Regression_Analysis_Marker}.

The QBC analysis confirms that input type is significant, with continuous input giving significantly greater AL performance than mixed; and mixed significantly greater than discrete.
%Thus the act of discretising the feature vectors, whether partially or completely, substantially reduces the AL performance.
This confirms that a discretised task is harder than a continuous one, with discretisation reducing the the diversity of pool examples. 
%Future work will examine that pool diversity in detail.

%REMOVED for new results with p2 and p10
%The QBC analysis also confirms that classifier optimum error rate is significant, increasing AL performance.

It is interesting that two very different AL methods lead to similar results for how AL performance depends on specific factors.
%It is noteworthy that two very different methods lead to similar results for how AL performance depends on specific factors.
We may explain this behaviour in part as follows.
With Active Learning there are two distinct stages: firstly the selection of examples for labelling, and secondly the use of those examples in training a particular classifier.
With SE the same classifier is used for both stages, whereas QBC uses a classifier committee for selection.
The QBC results found that classifier was not significant, in contrast to the SE results which found that Logistic Regression is significantly better than SVM.

This suggests that QBC may be selecting examples which are useful independently of the classifier: good datapoints which benefit any classifier.
That in itself is interesting, as it is a very plausible prior belief that the quality of datapoints would be strongly classifier dependent.

%This suggests that QBC may be selecting examples which are high quality independently of the classifier.
%That in itself is interesting, as it is widely believed in the literature that example quality is strongly classifier dependent (\cite{Wolpert1996}).

\section{Conclusion}
\label{Conclusion_Marker}

There are two central questions: Where does AL work? How much does it help?
By examining a variety of experiments across a range of points in AL factor space, some conclusions can be drawn.

Overall AL failed to demonstrate a performance gain far more often than not (11\% for SE, 6\% for QBC). 
This is consistent with several other authors who reported largely negative results using AL (\cite{Bach2006,Provost2010}).
The analysis also confirmed the general belief in the literature that AL provides its performance gain early on in the budget range.
Both AL methods, SE and QBC, showed that the smoothness of the input type makes a significant difference to AL performance.

In future we will extend this work, for example by including many more datasets, some from simulated data, other from real applications, e.g. \cite{Guyon2011}.
Future results should enable recommendations of AL method for applications, by relating the type of classification task to the relative performances of different AL methods.

This experimental study has generated some unexpected results about the factors that determine where AL works.
This study has shown many complexities with the assessment of AL performance.
It has contributed a new methodology to assess AL performance.

\subsection{Acknowledgement}

The work of Lewis P. G Evans is supported by an EPSRC doctoral training award.
%The authors would like to acknowledge the generosity of the EPSRC in funding this research.

%\bibliographystyle{plain}
%\bibliographystyle{plainnat}
%\bibliographystyle{ida_bib_sty}
%\bibliographystyle{IDA_llncs_v2}
%\bibliographystyle{doesn-not-exists_dasfgsdfsdfgsdfgsdfgsdfgdf_345645654645656}
%\bibliographystyle{ida_conf_2013_1}
%\bibliographystyle{plain_mod}
%\bibliography{llncs}

%\clearpage
%\addtocmark[2]{Author Index} % additional numbered TOC entry
%\renewcommand{\indexname}{Author Index}
%\printindex
%\clearpage
%\addtocmark[2]{Subject Index} % additional numbered TOC entry
%\markboth{Subject Index}{Subject Index}
%\renewcommand{\indexname}{Subject Index}
%\input{subjidx.ind}

\end{document}